\documentclass{article}

\usepackage{graphicx} 
\usepackage{epstopdf}
\usepackage{ifpdf}
\usepackage{subfigure} 
\usepackage{natbib}
\usepackage{algorithm}
\usepackage{algorithmic}
\usepackage{amsmath}
\usepackage{nicefrac}
\usepackage{amsfonts}
\usepackage{times}
\usepackage{amssymb}

\newcommand{\trueReturn}{\ensuremath{V}}
\newcommand{\empiricalReturn}{\ensuremath{V_{emp}}}

\newcommand{\episode}{\ensuremath{e}}
\newcommand{\policy}{\ensuremath{\pi}}

\newcommand{\policyClass}{\ensuremath{\Pi}}
\newcommand{\rewardFunction}{\ensuremath{\rho}}

\newcommand{\dynamics}{\ensuremath{m}}
\newcommand{\state}{\ensuremath{s}}
\newcommand{\stateSpace}{\ensuremath{{\cal S}}}

\newcommand{\startingState}{\ensuremath{\state_{start}}}
\newcommand{\action}{\ensuremath{a}}
\newcommand{\actionSpace}{\ensuremath{{\cal A}}}
\newcommand{\noise}{\ensuremath{w}}
\newcommand{\noiseSpace}{\ensuremath{{\cal W}}}
\newcommand{\episodeLength}{\ensuremath{T}}
\newcommand{\nEpisodes}{\ensuremath{N}}
\newcommand{\returnFunction}{\ensuremath{G}}

\newcommand{\inputVariable}{\ensuremath{x}}
\newcommand{\inputSpace}{\ensuremath{{\cal X}}}
\newcommand{\outputVariable}{\ensuremath{y}}
\newcommand{\outputSpace}{\ensuremath{{\cal Y}}}

\newcommand{\nDataPoints}{\ensuremath{N}}
\newcommand{\trueRisk}{\ensuremath{{\cal R}}}

\newcommand{\empiricalRisk}{\ensuremath{R_{emp}}}
\newcommand{\function}{\ensuremath{f}}
\newcommand{\functionEstimate}{\ensuremath{\hat{f}}}
\newcommand{\functionClass}{\ensuremath{{\cal F}}}

\newcommand{\loss}{\ensuremath{{\cal L}}}
\newcommand{\functionSet}{\ensuremath{S}}
\newcommand{\functionSetIndex}{\ensuremath{k}}

\newcommand{\lossMin}{\ensuremath{A}}
\newcommand{\lossMax}{\ensuremath{B}}
\newcommand{\VCDim}{\ensuremath{h}}

\newcommand{\rademacherVariable}{\ensuremath{R^\nDataPoints}}
\newcommand{\rademacherVariableEstimate}{\ensuremath{\hat{R}^\nDataPoints}}

\newcommand{\MFMCDistance}{\ensuremath{\Delta}}
\newcommand{\estimatedReturnMFMC}{\ensuremath{V_{MFMC}}}

\newcommand{\lipshitzConstant}{\ensuremath{L}}

\newcommand{\NewMFMCBoundConstant}{\ensuremath{d}}

\newcommand{\dataset}{\ensuremath{{\cal D}}}
\newcommand{\probOfFailure}{\ensuremath{\delta}}
\newcommand{\generalBoundConstant}{\ensuremath{\Omega}}

\newcommand{\IntruderDomainCameraRadiusActual}{0.5}
\newcommand{\IntruderDomainBadLocationMinDistActual}{0.05}
\newcommand{\IntruderDomainActionDomain}{\ensuremath{[ -0.1, 0.1 ]}}
\newcommand{\IntruderDomainBadLocationActual}{[0, 0]}
\newcommand{\IntruderDomainNumStatePointsActual}{16}
\newcommand{\IntruderDomainActionPointsActual}{two}

\newcommand{\IntruderDomainIntruderLocation}[1]{\ensuremath{X_{#1}}}
\newcommand{\IntruderDomainIntruderLocations}{\ensuremath{s_I}} 

\newcommand{\IntruderDomainWorldBounds}{\ensuremath{[[-1,1], [-1,1]]}}
\newcommand{\IntruderDomainCameraRadius}{\ensuremath{r_{\mathrm{cam}}}}
\newcommand{\IntruderDomainCameraLocation}{\ensuremath{s_{\textrm{cam}}}}

\newcommand{\IntruderDomainActionSymbol}{\ensuremath{a}}
\newcommand{\IntruderDomainCameraDynamics}{\ensuremath{\IntruderDomainCameraLocation{} = \IntruderDomainCameraLocation{} + \IntruderDomainActionSymbol{}}}
\newcommand{\IntruderDomainBadLocationSymbol}{\ensuremath{s_{\mathrm{sensitive}}}}
\newcommand{\IntruderDomainRewardFunction}{\ensuremath{ \rho(\IntruderDomainCameraLocation{},\IntruderDomainIntruderLocations{}) = \sum_i \frac{min( || \IntruderDomainCameraLocation{} - \IntruderDomainIntruderLocation{i} ||, \IntruderDomainCameraRadiusActual{} ) }{ max( || \IntruderDomainBadLocationSymbol{} - \IntruderDomainIntruderLocation{i} ||, \IntruderDomainBadLocationMinDistActual{} ) } }}
\newcommand{\RadiaBasisFunctionWeightI}{\ensuremath{\phi_{i}}}

\title{\bf Structural Return Maximization for Reinforcement Learning}
\author{Joshua Joseph, Javier Velez, Nicholas Roy \\ ~ \\ 
{\it Massachusetts Institute of Technology} \\
\texttt{\{jmjoseph, velezj, nickroy\}@mit.edu}}

\begin{document} 

\maketitle

\begin{abstract} 

Batch Reinforcement Learning (RL) algorithms attempt to choose a policy from a designer-provided class of policies given a fixed set of training data.  
Choosing the policy which maximizes an estimate of return often leads to over-fitting when only limited data is available, due to the size of the policy class in relation to the amount of data available.
In this work, we focus on learning policy classes that are appropriately sized to the amount of data available.
We accomplish this by using the principle of Structural Risk Minimization, from Statistical Learning Theory, which uses Rademacher complexity to identify a policy class that maximizes a bound on the return of the best policy in the chosen policy class, given the available data.
Unlike similar batch RL approaches, our bound on return requires only extremely weak assumptions on the true system.
\end{abstract} 

\section{Introduction}
\label{sec:introduction}
Reinforcement Learning (RL) \citep{sutton98reinforcement} is a
framework for sequential decision making under uncertainty with the
objective of finding a policy that maximizes the sum of rewards, or
{\it return}, of an agent. A straightforward model-based approach to batch RL,
where the algorithm learns a policy from a fixed set of data, is to
fit a dynamics model by minimizing a form of prediction error ({\it
  e.g.}, minimum squared error) and then compute the optimal policy
with respect to the learned model \citep{bertsekas00dynamic}.  As
discussed in \citet{baxter01infinite} and \citet{joseph13reinforcement}, learning a model for RL by
minimizing prediction error can result in a policy that performs
arbitrarily poorly for unfavorably chosen model classes.  
To overcome this limitation, a second approach is to not use a model and directly learn
the policy from a policy class that explicitly maximizes an estimate of return
\citep{meuleau00offpolicy}.

With limited data, approaches that explicitly maximize estimated return are vulnerable to learning policies
which perform poorly since the return cannot be confidently
estimated.  We overcome this problem by applying the principle
of Structural Risk Minimization (SRM) \citep{vapnik98statistical},
which, in terms of RL, states that instead of choosing the policy which
maximizes the estimated return we should instead maximize a bound on
return.  In SRM the policy
class size is treated as a controlling variable in the optimization of the bound,
allowing us to naturally trade-off between estimated performance and
estimation confidence.
By controlling policy class size in this principled way we can overcome the poor performance of approaches which explicitly maximize estimated return with small amounts of data.

The main contribution of this work is a batch RL algorithm which has bounded true return under extremely weak assumptions, unlike standard batch RL approaches.
Our algorithm is the result of applying the principle of SRM to
RL, which previously has only been studied in the context of
classification and regression.
We first show a bound on the return of a single policy from a fixed policy class based on a technique called Model-Free Monte Carlo \citep{fonteneau12batch}.
We then map RL to classification, allowing us to
transfer generalization bounds based on
Rademacher complexity \citep{bartlett03rademacher} which results in a
bound on the return of any policy from a policy class.  Given a
structure of policy classes, we then apply the principle of SRM to
find the highest performing policy from the family of policy classes.

Section \ref{sec:srm_classification} reviews Structural Risk
Minimization in the context of classification.  We move to RL in
Section \ref{sec:policy_return_bound} and show the bound on the return of a policy.
Section \ref{sec:srm} ties together the previous two sections to
provide a bound on return for a policy from a structure of policy
classes and discusses some of the natural policy class
structures that exist in RL.
Section \ref{sec:results} first demonstrates our approach on a simple domain
to build intuition for the reader and then validates its performance on
increasingly difficult problems.  Sections \ref{sec:related_work}
discusses related work and Section \ref{sec:conlusion} concludes the
paper.

\section{Structural Risk Minimization for Classifier Learning}
\label{sec:srm_classification}
In this section we review Structural Risk Minimization for classification for completeness and to ensure that the parallels presented in Section \ref{sec:srm} are clear for the reader.
Classification is the problem of deciding on an output, $\outputVariable \in \outputSpace$, for a given input, $\inputVariable \in \inputSpace$.
The performance of a decision rule ${\function:\inputSpace \rightarrow \outputSpace}$ is measured using risk, $\trueRisk$, defined as
\begin{equation}
\trueRisk(\function) = \int \loss(\outputVariable, \function(\inputVariable)) p(\inputVariable, \outputVariable) d \inputVariable ~ d \outputVariable \label{eq:risk}
\end{equation}
where $\loss : \outputSpace \times \outputSpace \rightarrow \mathbb{R}$ is the loss function and $p(\inputVariable, \outputVariable)$ is the data generating distribution.
For a class of decision rules, $\functionClass$, the objective of classification is to select the decision rule which minimizes risk or, more formally,
\begin{equation}
\function^* = \arg \min_{\function \in \functionClass} \trueRisk(\function). \label{eq:classification_objective}
\end{equation}

\subsection{Empirical Risk Minimization}
Commonly, the distribution $p(\inputVariable, \outputVariable)$ in Equation \ref{eq:risk} is unknown, and we therefore are unable solve Equation \ref{eq:classification_objective} using Equation \ref{eq:risk}.
Given a dataset 
$\dataset = \{(\inputVariable^1,\outputVariable^1), (\inputVariable^2,\outputVariable^2), \dots, (\inputVariable^\nDataPoints,\outputVariable^\nDataPoints)\}$
where $(\inputVariable^n,\outputVariable^n)$ is drawn i.i.d. from $p(\inputVariable, \outputVariable)$, Equation \ref{eq:risk} can be approximated by empirical risk
\begin{equation}
\empiricalRisk(\function; \dataset) = \frac{1}{\nDataPoints} \sum_{n=1}^\nDataPoints \loss(\outputVariable^n, \function(\inputVariable^n)) . \label{eq:empirical_risk}
\end{equation}
By using empirical risk as an estimate of risk we can attempt to solve Equation \ref{eq:classification_objective} using the principle of Empirical Risk Minimization (ERM) \citep{vapnik98statistical} where
\begin{equation}
\functionEstimate = \arg \min_{\function \in \functionClass} \empiricalRisk(\function; \dataset). \label{eq:empirical_risk_minimization}
\end{equation}
In Section \ref{sec:policy_return_bound} we see that there is a clear analogy between this result and how a policy's return is estimated in RL.

\subsection{Bounding the Risk of a Classifier}
\label{sec:classifier_bound}
We can bound risk (Equation \ref{eq:risk}) using using empirical risk (Equation \ref{eq:empirical_risk}) with a straightforward application of Hoeffding's inequality
\begin{equation}
\trueRisk(\function) \leq \empiricalRisk(\function; \dataset) + \sqrt{\frac{-\ln \delta}{2 \nDataPoints}} \label{eq:empirical_estimation_error}
\end{equation}
which holds with probability $1-\probOfFailure$.
Since Equation \ref{eq:empirical_risk_minimization} is used to choose $\function \in \functionClass$, we need to ensure that $\trueRisk(\function)$ is bounded for {\it all} $\function \in \functionClass$ (not just for a single $f$ as Equation \ref{eq:empirical_estimation_error} guarantees).
Bounds of this form ($\forall \function \in \functionClass$) can be written as
\begin{equation}
\trueRisk(\function) \leq \empiricalRisk(\function; \dataset) + \generalBoundConstant(\loss \circ \functionClass, \dataset, \probOfFailure) \label{eq:risk_bound}
\end{equation}
where \generalBoundConstant{} can be thought of as a complexity penalty on the size of $\loss \circ \functionClass = \{\loss(\cdot, \function(\cdot)) : \function \in \functionClass\}$ and the bound holds with probability $1-\probOfFailure$.

Section \ref{sec:vc_bound} and \ref{sec:rademacher_bound} describes a specific forms of $\generalBoundConstant$ using Vapnik-Chervonenkis Dimension and Rademacher complexity, which we chose due to their popularity in the literature although many additional bounds are known, {\it e.g.}, maximum discrepancy \citep{bartlett02model}, local Rademacher complexity \citep{bartlett02local}, Gaussian complexity \citep{bartlett03rademacher}.

\subsubsection{Vapnik-Chervonenkis Dimension}
\label{sec:vc_bound}
A well studied bound from \citet{vapnik95thenature} that takes the form of Equation \ref{eq:risk_bound} uses
\begin{equation}
\generalBoundConstant(\loss \circ \functionClass, \dataset, \probOfFailure) = \frac{\lossMax - \lossMin}{2} \sqrt{4\frac{\VCDim \left ( \ln \left ( \frac{2 \nDataPoints}{\VCDim} \right ) + 1 \right ) - \ln(\delta/4)}{\nDataPoints}} \label{eq:vc_bound}
\end{equation}
where $\lossMin \leq \loss \circ \functionClass \leq \lossMax$ and $\VCDim$ is the Vapnik-Chervonenkis (VC) dimension of $\loss \circ \functionClass$ (see \citet{vapnik98statistical} for a thorough description of VC dimension).

\subsubsection{Rademacher Complexity}
\label{sec:rademacher_bound}
In contrast to the well studied bound from \citet{vapnik95thenature}
which depends on the Vapnik-Chervonenkis (VC) dimension of $\loss
\circ \functionClass$,  \citet{bartlett03rademacher} provide a bound
based on the Rademacher complexity of $\loss \circ \functionClass$, a
quantity that can be straightforwardly estimated from the dataset.
Their bound, which takes the form of Equation \ref{eq:risk_bound},
uses
\begin{equation}
\generalBoundConstant(\loss \circ \functionClass, \dataset, \probOfFailure) = \rademacherVariable(\loss \circ \functionClass) + \sqrt{\frac{-8 \ln (2 \probOfFailure)}{\nDataPoints}} \label{eq:rademacher_bound}
\end{equation}
where $0 \leq \loss \circ \functionClass \leq 1$, and $\sigma_n$ is a uniform random variable over $\{-1, +1\}$. 
$\rademacherVariable(\functionClass)$, the Rademacher complexity of $\functionClass$, can be estimated using 
\begin{align}
\rademacherVariable(\functionClass) & = E_{\sigma_{1:\nDataPoints},\inputVariable^{1:\nDataPoints}} \left [ \sup_{\function \in \functionClass} ~ \frac{2}{\nDataPoints} \left | \sum_{n=1}^\nDataPoints \sigma_n \function(\inputVariable^n) \right | \right ], \nonumber \\
\rademacherVariableEstimate(\functionClass; \dataset) & = E_{\sigma_{1:\nDataPoints}} \left [ \left. \sup_{\function \in \functionClass} ~ \frac{2}{\nDataPoints} \left | \sum_{n=1}^\nDataPoints \sigma_n \function(\inputVariable^n) \right | ~ \right | \inputVariable^{1:\nDataPoints} \right ]. \label{eq:rademacher_estimate} 
\end{align}
\citet{bartlett03rademacher} also show that the error from estimating Rademacher complexity using the right hand side of Equation \ref{eq:rademacher_estimate} is bounded with probability $1-\delta$ by
\begin{equation}
\rademacherVariable(\functionClass) \leq \rademacherVariableEstimate(\functionClass; \dataset) + \sqrt{\frac{-8 \ln \delta}{\nDataPoints}} . \label{eq:rademacher_estimate_bound} 
\end{equation}

\subsection{Structural Risk Minimization}
\label{sec:srm_subsection}
As discussed in \citet{vapnik95thenature}, the principle of ERM is only intended to be used with a large amount of data (relative to the size of $\functionClass$).
With a small data set, a small value of $\empiricalRisk(\function; \dataset)$ does not guarantee that $\trueRisk(\function)$ will be small and therefore solving Equation \ref{eq:empirical_risk_minimization} says little about the generalization of $\function$.
The principle of Structural Risk Minimization (SRM) states that since we cannot guarantee the generalization of ERM under limited data we should explicitly minimize the bound on generalization (Equation \ref{eq:risk_bound}) by using a {\it structure} of function classes.

A structure of function classes is defined as a collection of nested subsets of functions ${\functionSet_1 \subseteq \functionSet_2 \subseteq \cdots \subseteq \functionSet_\functionSetIndex \subseteq \cdots}$
where $\functionSet_\functionSetIndex = \loss \circ \functionClass_\functionSetIndex$.
For example, a structure of radial basis functions created by placing increasing limits on the magnitude of the basis functions.
SRM then treats the capacity of $\loss \circ \functionClass_\functionSetIndex$ as a controlling variable and minimizes Equation \ref{eq:risk_bound} for each $\functionSet_\functionSetIndex$ such that
\begin{align}
\hat{\functionSetIndex} & = \arg \min_{\functionSetIndex} \empiricalRisk(\hat{\function}_\functionSetIndex; \dataset) + \generalBoundConstant(\loss \circ \functionClass_\functionSetIndex, \dataset, \probOfFailure) \nonumber \\
\hat{\function}_\functionSetIndex & = \arg \min_{\function \in \functionClass_\functionSetIndex} \empiricalRisk(\function; \dataset). \label{eq:srm_objective}
\end{align}
To solve Equation \ref{eq:srm_objective} we must solve both equations jointly.
One can imagine enumerating $k$, finding $\hat{\function}_\functionSetIndex$ for each $k$, and choosing the corresponding $\hat{\function}_\functionSetIndex$ which minimizes $\empiricalRisk(\hat{\function}_\functionSetIndex; \dataset) + \generalBoundConstant(\loss \circ \functionClass_\functionSetIndex, \dataset, \probOfFailure)$.

\section{Bounding the Return of a Policy}
\label{sec:policy_return_bound}
\label{sec:rl}
Section \ref{sec:srm_classification} allowed us to bound classification performance given small amounts of data; we now turn out attention to bounding policy performance.
A finite time Markov Decision Process (MDP) is defined as a tuple $(\stateSpace, \actionSpace, \noiseSpace, \dynamics, \rewardFunction, \startingState, \episodeLength)$ where \stateSpace{} is the state space, \actionSpace{} is the action space, \noiseSpace{} is the disturbance space\footnote{The disturbance space is introduced so we may assume the dynamics model is deterministic and add noise through $\noise$. This is primarily done to facilitate theoretical analysis and is equivalent to the standard RL notation which uses a stochastic dynamics model that does not include $\noise$.},
${\dynamics : \stateSpace \times \actionSpace \times \noiseSpace \rightarrow \stateSpace}$ is the dynamics model, ${\rewardFunction : \stateSpace \rightarrow \mathbb{R}}$ is the reward function, $\startingState \in \stateSpace$ is the starting state, and \episodeLength{} is the maximum length of an episode\footnote{For simplicity we assume \rewardFunction{} and $\state_0 = \startingState{}$ are known and deterministic and \rewardFunction{} is only a function of the current state. 
Without loss of generality this allows us to write $\trueReturn$ as a scalar everywhere.
This work can be straightforwardly extended an unknown and stochastic $\rewardFunction$ and $\startingState{}$ and the more general ${\rewardFunction : \stateSpace \times \actionSpace \times \stateSpace \rightarrow \mathbb{R}}$.}.
For a policy, $\policy : \stateSpace \rightarrow \actionSpace$, we define its return\footnote{$\trueReturn(\policy)$ is shorthand for $V^{\policy}(\startingState)$, the expected sum of rewards from following policy $\pi$ from state $\startingState{}$.}
as
\begin{align}
& \trueReturn(\policy) = \int_{\noise_{0:T-1}} \Bigg [ \rewardFunction(\state_0) ~ + \nonumber \\
& ~~~~~~~~ \sum_{t=0}^{T-1} \rewardFunction(\dynamics(\state_{t}, \policy(\state_{t}), \noise_{t})) \Bigg ] ~ p(\noise_{0:T-1}) ~ d \noise_{0:T-1} \label{eq:true_return_1} \\
& = E_{\noise_{0:T-1}} \Bigg [ \sum_{t=0}^{\episodeLength-1} \rewardFunction(\state_t) \Bigg | \state_0 = \startingState, \nonumber \\ 
&~~~~~~~~~~~~~~~~~~~~~~~~~~~~~~~~~~~~~~~\state_{t+1} = \dynamics(\state_t, \policy(\state_t), \noise_t) \Bigg ] .  \label{eq:true_return}
\end{align}
We call the sequence ${\episode_\policy = \{\state_0, \action_0, \state_1, \action_1, \dots, \state_{\episodeLength-1}, \action_{\episodeLength-1}\}}$
an episode of data where $\state_{t+1} = \dynamics(\state_t, \action_t, \noise_t)$ and $\action_t = \policy(\state_t)$.

Throughout this work we assume that the return is bounded by $\lossMin \leq \sum_{t=0}^{\episodeLength-1} \rewardFunction(\state_t) \leq \lossMax$ and that the dynamics model, reward function, and policy are Lipschitz continuous with constants $\lipshitzConstant_\dynamics, \lipshitzConstant_\rewardFunction$, and $\lipshitzConstant_\policy$, respectively \citep{fonteneau12batch}, where
$$
||\dynamics(\state,\action,\noise)-\dynamics(\state',\action',\noise)||_\stateSpace \leq \lipshitzConstant_\dynamics (||\state-\state'||_\stateSpace + ||\action-\action'||_\actionSpace),
$$
$$
|\rewardFunction(\state,\action,\noise)-\rewardFunction(\state',\action',\noise)| \leq \lipshitzConstant_\rewardFunction (||\state-\state'||_\stateSpace + ||\action-\action'||_\actionSpace),
$$
$$
||\policy(t,\state)-\policy(t,\state')||_\actionSpace \leq \lipshitzConstant_\policy ||\state-\state'||_\stateSpace,
$$
$$
\forall (\state,\state',\action,\action',\noise) \in \stateSpace^2 \times \actionSpace^2 \times \noiseSpace,~\forall t \in \{0,..,\episodeLength-1\}.
$$
The objective of an MDP is to find the policy 
\begin{equation}
\policy^* = \arg \max_{\policy \in \policyClass} \trueReturn(\policy) \label{eq:rl_objective}
\end{equation}
from a given policy class, $\policyClass$. 
Typically in RL the dynamics model, \dynamics{}, is unavailable to us and therefore we are unable use Equation \ref{eq:true_return} to solve Equation \ref{eq:rl_objective}.

\subsection{Estimating the Return of a Policy}
\label{sec:estimating_return}
To overcome not knowing \dynamics{} in Equation \ref{eq:true_return} we commonly use data of interactions with \dynamics{} to estimate $\trueReturn(\policy)$, called policy evaluation, and then solve Equation \ref{eq:rl_objective}.
A difficulty in estimating $\trueReturn(\policy)$ lies in how the data was collected, or, more precisely, which policy was used to generate the data.
We discuss two different types of policy evaluation, {\it on-policy}, where the data used to estimate $\trueReturn(\policy)$ is generated using $\policy$, and {\it off-policy}, where the data is collected using any policy.

\subsubsection{On-Policy Policy Evaluation}
\label{sec:on_policy}
Given a set of $\nEpisodes$ episodes of data $\{\episode_\policy^1,
\episode_\policy^2, \dots, \episode_\policy^\nEpisodes\}$ collected
using $\policy$, we can estimate $\trueReturn(\policy)$ using
empirical return (analogous to Equation \ref{eq:empirical_risk}) where
\begin{align}
& \empiricalReturn(\policy; \dataset) = \frac{1}{\nEpisodes} \sum_{n=1}^\nEpisodes \sum_{t=0}^{\episodeLength-1} \rewardFunction(\state^n_t) \label{eq:empirical_return} , \\
& \trueReturn(\policy) \geq \empiricalReturn(\policy; \dataset) - (\lossMax-\lossMin) \sqrt{\frac{-\ln \probOfFailure}{2 \nEpisodes}} , \label{eq:hoeffding}
\end{align}
$\episode_\policy^n = \{\state_0^n, \action_0^n, \dots,
\state_{\episodeLength-1}^n, \action_{\episodeLength-1}^n\}$,
$\state_{t+1}^n = \dynamics(\state_t^n, \action_t^n, \noise_t^n)$,
$\action_t^n = \policy(\state_t^n)$, and Equation \ref{eq:hoeffding}
holds with probability $1-\probOfFailure$ by Hoeffding's inequality
\citep{hoeffding63probability}.  This approach, where episodes are
generated on-policy and then the return is estimated using Equation
\ref{eq:empirical_return}, is called Monte Carlo policy evaluation
\citep{sutton98reinforcement}.
Since we do not make any assumptions about the policies that will be evaluated nor the policy under which the data was generated, we cannot directly use Equations \ref{eq:empirical_return} and \ref{eq:hoeffding} but we will build upon them in the following sections.


\subsubsection{Off-Policy Policy Evaluation} 
\label{sec:mfmc_pac_bound}

Naively, using Equation \ref{eq:empirical_return} to approximate and solve Equation \ref{eq:rl_objective}, would require $\nEpisodes$ data for each $\policy \in \policyClass$, an infinite amount of data for infinite policy classes.
Off-policy policy evaluation aims to alleviate this issue by estimating $\trueReturn(\policy)$ using episodes $\episode_{\policy_1}^1, \episode_{\policy_2}^2, \dots, \episode_{\policy_\nEpisodes}^\nEpisodes$ where $\policy_1, \dots, \policy_\nEpisodes$ may be different than $\policy$.
To perform off-policy evaluation, we use an approach called Model-Free Monte Carlo-like policy evaluation (MFMC) \citep{fonteneau12batch}, which attempts to approximate the estimator from Equation \ref{eq:empirical_return} by piecing together artificial episodes of on-policy data from off-policy, batch data.

Consider a set of data $\{\episode_{\policy_1}^1, \episode_{\policy_2}^2, \dots, \episode_{\policy_\nEpisodes}^\nEpisodes\}$, which we re-index as one-step transitions 
$$
{\dataset = \{(\state_0, \action_0, \state_1), \dots, (\state_{NT-1}, \action_{NT-1}, \state_{NT})\}}.
$$
To evaluate a policy, $\policy$, MFMC uses a distance function 
$$
\MFMCDistance((\state, \action), (\state', \action')) = || \state-\state'||_\stateSpace + || \action-\action' ||_\actionSpace
$$ 
and pieces together $\tilde{\nEpisodes}$ artificial episodes from $\dataset$ such that $\tilde{\episode}^n_\policy = \{\tilde{\state}_0^n, \tilde{\action}_0^n, \dots, \tilde{\state}_T^n, \tilde{\action}_T^n \}$ is an artificial on-policy episode approximating an episode $\policy$ for ${n = 1, \dots, \tilde{\nEpisodes}}$.
To construct $\tilde{\episode}^n_\policy$, MFMC starts with $\tilde{\state}_0^n = \startingState$ and for ${t = 1, \dots, \episodeLength}$ we find
\begin{equation}
\tilde{\state}_{t+1}^n = \arg \min_{s' : (s,a,s') \in \dataset} \MFMCDistance((\state, \action), (\tilde{\state}_t^n, \tilde{\action}_t^n)) \label{eq:mfmc_next_state}
\end{equation}
where $\tilde{\action}_t^n = \policy(\tilde{\state}_t^n)$ and once a transition, $(s,a,s')$, is chosen using Equation \ref{eq:mfmc_next_state} it is removed from \dataset{}.
Following the construction of episodes $\tilde{\episode}^1_\policy, \dots, \tilde{\episode}^{\tilde{\nEpisodes}}_\policy$, MFMC estimates the return of $\policy$ using
\begin{equation}
\estimatedReturnMFMC(\policy; \dataset) = \frac{1}{\tilde{\nEpisodes}} \sum_{n=1}^{\tilde{\nEpisodes}} \sum_{t=0}^{\episodeLength-1} \rewardFunction(\tilde{\state}^n_t) . \label{eq:tmp_1}
\end{equation}

We can bound the return using Theorem 4.1, Lemma A.1, and Lemma A.2 of \citet{fonteneau10model} and say that 
\begin{equation}
\trueReturn(\policy) \geq E_{\dataset} \left [ \estimatedReturnMFMC(\policy) \right ] - \NewMFMCBoundConstant(\policy; \dataset) \label{eq:mfmc_expectation_bound} 
\end{equation}
where 
\begin{align}
& \NewMFMCBoundConstant(\policy; \dataset) = \max_{n \leq \tilde{N}} \sum_{t=0}^{\episodeLength-1} L_{\episodeLength-t} \delta_t^n, \nonumber \\
& \delta_t^n = \min_{s' : (s,a,s') \in \dataset} \MFMCDistance((\state, \action), (\tilde{\state}_t^n, \tilde{\action}_t^n)) \nonumber
\end{align} 
for each $\tilde{\state}_{t+1}^n$ chosen using Equation \ref{eq:mfmc_next_state}, and 
$$
L_{\episodeLength-t} = \lipshitzConstant_\rewardFunction \sum_{i=0}^{\episodeLength-t-1} [\lipshitzConstant_\dynamics(1+\lipshitzConstant_\policy)]^i.
$$
The term $\NewMFMCBoundConstant(\policy; \dataset)$ is the maximum deviation between the true return of any policy and the expected MFMC estimate of return of that policy.
See \citet{fonteneau10model, fonteneau12batch} for a discussion regarding the choice of $\MFMCDistance$ and $\tilde{\nEpisodes}$.

\subsubsection{Probabilistic Bound of the MFMC Estimator}
\label{sec:prob_mfmc_est}
Unfortunately, the bound provided in Equation \ref{eq:mfmc_expectation_bound} only allows us to bound the return using the {\it expectation} of the MFMC estimate, not the {\it realized} estimate based on the data, which is required in Section \ref{sec:srm}.
We present such a bound, beginning with Hoeffding's inequality \citep{hoeffding63probability}\footnote{Note MFMC only allows a single transition to be used once in constructing a set of episodes. Therefore, the return from the episodes do not violate the independence assumption required to use Hoeffding's inequality.},
\begin{align}
\lefteqn{e^{2 \nEpisodes \left ( \epsilon - \NewMFMCBoundConstant(\policy; \dataset) \right )^2/(\lossMax-\lossMin)^2}} \nonumber \\
& > \mathrm{Pr}\left [ \estimatedReturnMFMC(\policy; \dataset) - E [ \estimatedReturnMFMC(\policy) ] + \NewMFMCBoundConstant(\policy; \dataset) > \epsilon \right ] \label{eq:mfmc_pac_proof_1} \\
& \geq \mathrm{Pr}\left [ \estimatedReturnMFMC(\policy; \dataset) - \trueReturn(\policy) > \epsilon \right ] \label{eq:mfmc_pac_proof_2}
\end{align}
where we move from Equation \ref{eq:mfmc_pac_proof_1} to Equation \ref{eq:mfmc_pac_proof_2} using Equation \ref{eq:mfmc_expectation_bound}.
Setting 
$\probOfFailure = e^{2 \nEpisodes \left ( \epsilon - \NewMFMCBoundConstant(\policy; \dataset) \right )^2/(\lossMax-\lossMin)^2}$, solving for $\epsilon$, and substituting the quantity into Equation \ref{eq:mfmc_pac_proof_2} we see that with at least probability $1-\probOfFailure$

\begin{equation}
\trueReturn(\policy) \geq \estimatedReturnMFMC(\policy; \dataset) - \NewMFMCBoundConstant(\policy; \dataset) - (\lossMax-\lossMin) \sqrt{\frac{-\ln \probOfFailure}{2 \nEpisodes}}. \label{eq:mfmc_pac_bound}
\end{equation}
While Equation \ref{eq:mfmc_pac_bound} is useful for bounding the return estimate using MFMC, in Section \ref{sec:srm} we will require a bound between $\estimatedReturnMFMC(\policy; \dataset)$ and $\empiricalReturn(\policy)$.
By combining Equations \ref{eq:hoeffding} and \ref{eq:mfmc_pac_bound}, we have with probability $1-2 \probOfFailure$
\begin{align}
\empiricalReturn(\policy; \dataset) \geq & ~ \estimatedReturnMFMC(\policy; \dataset) - \NewMFMCBoundConstant(\policy; \dataset) \nonumber \\ 
& ~~~~~~ - 2 (\lossMax-\lossMin) \sqrt{\frac{-\ln \probOfFailure}{2 \nEpisodes}}. \label{eq:mfmc_pac_bound_to_estimate_return}
\end{align}

\section{Structural Return Maximization for Policy Learning}
\label{sec:srm}

Sections \ref{sec:srm_classification} and \ref{sec:policy_return_bound} provide intuition for the similarities between classification and RL, {\it e.g.}, in classification we choose $\function \in \functionClass$ using Equations \ref{eq:risk} and \ref{eq:classification_objective}, and in RL we choose $\policy \in \policyClass$ using Equations \ref{eq:true_return} and \ref{eq:rl_objective}.
In this section we aim to formalize the relationship between classification and RL and by doing so we are able to use Structural Risk Minimization (Section \ref{sec:srm_subsection}) to learn policies drawn from policy classes that are appropriately sized given the available data.

\subsection{Mapping Classification to Reinforcement Learning}

The difficulty in mapping classification to RL is most easily seen when we consider the bounds in Sections \ref{sec:vc_bound} and \ref{sec:rademacher_bound}, for which we need to know the RL equivalent of $\loss \circ \functionClass$ to either compute its VC dimension (for Equation \ref{eq:vc_bound}) or its Rademacher complexity (for Equation \ref{eq:rademacher_bound}).
To show the mapping we begin by defining the {\it return function},
\begin{equation}
\returnFunction(\state_0, \noise_{0:\episodeLength-1}, \action_{0:\episodeLength-1}) \triangleq \rewardFunction(\state_0) + \sum_{t=0}^{\episodeLength-1} \rewardFunction(\dynamics(\state_{t}, \action_{t}, \noise_{t})) , \label{eq:return_function}
\end{equation}
and show that the classification objective of minimizing risk is equivalent to the RL objective of maximizing return.
Using Equation \ref{eq:classification_objective} and a loss function that does not depend on $\outputVariable$\footnote{While it may seem that writing $\loss$ without $\outputVariable$ is an abuse of notation, if instead we view $\loss$ as a measure of performance the relationship between RL and classification becomes more clear.},
we set $\loss(\outputVariable, \function(\inputVariable)) = \tilde{\loss}(\inputVariable, \function)$ and see that
\begin{align}
\trueRisk(\function) & = \int \tilde{\loss}(\inputVariable, \function)~ p(\inputVariable) ~ d \inputVariable \label{eq:mapping_1} \\
& = \int -\tilde{\returnFunction}(\state_0, \noise_{0:\episodeLength-1}, \function)~ p(\state_0, \noise_{0:\episodeLength-1}) ~ d \noise_{0:\episodeLength-1}  \label{eq:mapping_2} \\
& = - \trueReturn(\function) \label{eq:mapping_3}
\end{align}
where we go from Equation \ref{eq:mapping_1} to Equation \ref{eq:mapping_2} by setting ${\inputVariable = [\state_0, \noise_{0:\episodeLength-1}]}$ and noting that ${\tilde{\loss}(\inputVariable, \function) = -\tilde{\returnFunction}(\state_0, \noise_{0:\episodeLength-1}, \function) = -\returnFunction(\state_0, \noise_{0:\episodeLength-1}, \action_{0:\episodeLength-1})}$ for ${\action_t = \function(\state_t)}$, and we move from Equation \ref{eq:mapping_2} to Equation \ref{eq:mapping_3} using Equation \ref{eq:true_return}.
Therefore, minimizing $\trueRisk(\function)$ for $\function \in \functionClass$ is identical to maximizing $\trueReturn(\function)$ for $\function \in \functionClass$.
We see that $\returnFunction$ is the term in brackets in Equation~\ref{eq:true_return_1}; $\returnFunction$ encodes both the reward function and dynamics model and is analogous to classification's $\loss$.
This is a crucial relationship since transferring the bounds from Section \ref{sec:classifier_bound} depends on our being able to calculate the RL equivalent of $\loss \circ \functionClass$, which we denote $\returnFunction \circ \policyClass$\footnote{Note that it is difficult to tell how deep the connection between classification and RL is and, while we think this will prove an interesting line of future research, the purpose of this connection is solely to provide us with the machinery necessary for the bounds.}.

\subsection{Bound on Return for all Policies in a Policy Class}
Using the mapping from the previous section we can rewrite Equations \ref{eq:empirical_risk} and \ref{eq:risk_bound} for RL as 
\begin{align}
\empiricalReturn(\policy; \dataset) & = \frac{1}{\nEpisodes} \sum_{n=1}^\nEpisodes G(\state_0, \noise^{\policy,n}_{0:\episodeLength-1},\action^{\policy,n}_{0:\episodeLength-1}) \nonumber \\ 
& = \frac{1}{\nEpisodes} \sum_{n=1}^\nEpisodes \sum_{t=0}^\episodeLength \rewardFunction(\state^{\policy,n}_t) \label{eq:empirical_return_1} \\
\trueReturn(\policy) & \geq \empiricalReturn(\policy; \dataset) - \generalBoundConstant(\returnFunction \circ \policyClass, \dataset, \probOfFailure) . \label{eq:rl_bound}
\end{align}
where $\state^{\policy,n}_{t+1} = \dynamics((\state^{\policy,n}_t, \policy(\state^{\policy,n}_t), \noise^n_{0:\episodeLength-1})$.
Note that in Equation \ref{eq:empirical_return_1} $\noise^n_{0:\episodeLength-1}$ is needed to compute the empirical return which typically is not observed in practice.
Since we assume the state is fully observable, we can use Equation \ref{eq:empirical_return_1} to calculate empirical return.
Combining Equations \ref{eq:mfmc_pac_bound_to_estimate_return} and \ref{eq:rl_bound}, 
\begin{align}
\trueReturn(\policy) \geq & \estimatedReturnMFMC(\policy; \dataset) - \NewMFMCBoundConstant(\policy; \dataset) \nonumber \\ 
& ~~~~ - 2 (\lossMax-\lossMin) \sqrt{\frac{-\ln \probOfFailure}{2 \nEpisodes}} - \generalBoundConstant(\returnFunction \circ \policyClass,\dataset, \probOfFailure) \label{eq:combined_1}
\end{align}
which holds with probability $1-3 \probOfFailure$ and is the bound on return for all policies in $\policyClass$.
Section \ref{sec:rademacher_rl_bound} describes the use Rademacher complexity (Section \ref{sec:rademacher_bound}) to compute $\generalBoundConstant$ in Equation \ref{eq:combined_1}.

\subsection{Bound Based on VC Dimension for RL}
\label{sec:vc_bound_rl}
To use the bound described in Section \ref{sec:vc_bound} we need to know the VC dimension of $G \circ \policyClass$.
Unfortunately, the VC dimension is only known for specific function classes ({\it e.g.}, linear indicator functions \citep{vapnik98statistical}), and, since the only assumptions we made about functional form of $\rewardFunction$, $\dynamics$, and $\policy$ is that they are Lipshitz continuous, $G \circ \policyClass$ will not in general have known VC dimension.

There also exist known bounds on the VC dimension for {\it e.g.}, neural networks \citep{anthony99neural}, decision trees \citep{asian09calculating}, support vector machines \citep{vapnik98statistical}, smoothly parameterized function classes \citep{lee95lower}, but, again, due to our relatively few assumptions on $\rewardFunction$, $\dynamics$, and $\policy$, $G \circ \policyClass$ is not a function class with a known bound on the VC dimension.
\citet{cherkassky98learning} notes that for some classification problems, the VC dimension of $\functionClass$ can be used as a reasonable approximation for the VC dimension $\loss \circ \functionClass$, but we have no evidence to support this being an accurate approximation when used in RL.

Other work has been done to estimate the VC dimension from data \citep{shao69measuring, vapnik94measuring} and bound the VC dimension estimate \citep{mcdonald11estimated}.
While, in principle, we are able to estimate the VC dimension using one of these techniques, the approach described in Section \ref{sec:rademacher_rl_bound} is a far simpler method for computing $\Omega$ in Equation \ref{eq:rl_bound} based on data.

\subsection{Bound Based on Rademacher Complexity for RL}
\label{sec:rademacher_rl_bound}

Using the Rademacher complexity bound (Section \ref{sec:rademacher_bound}) allows us to calculate $\Omega$ (Equation \ref{eq:rl_bound}) based on data.
The only remaining piece is how to calculate the summation inside the absolute value sign of Equation \ref{eq:rademacher_estimate} for RL.
Mapping the Rademacher complexity estimator (Equation \ref{eq:rademacher_estimate}) into RL yields
\begin{equation}
\rademacherVariableEstimate(G \circ \policyClass; \dataset) = E_{\sigma_{1:\tilde{\nEpisodes}}} \Bigg [\sup_{\policy \in \policyClass} ~ \frac{2}{\tilde{\nEpisodes}} \Bigg | \sum_{n=1}^{\tilde{\nEpisodes}} \sigma_n \sum_{t=0}^{\episodeLength-1} \rewardFunction(\state_t^{n}) \Bigg | ~ \Bigg | \dataset \Bigg ]. \label{eq:rademacher_estimate_mfmc_1}
\end{equation}
Therefore, with probability $1-\probOfFailure$
\begin{align}
\lefteqn{\rademacherVariableEstimate(G \circ \policyClass; \dataset) \geq} \nonumber \\
& ~~~~~ - 2 + E_{\sigma_{1:\tilde{\nEpisodes}}} \Bigg [ \sup_{\policy \in \policyClass} ~ \frac{2}{\tilde{\nEpisodes}} \Bigg | \sum_{n=1}^{\tilde{\nEpisodes}} \sigma_n \Bigg ( \estimatedReturnMFMC(\policy; \dataset) \nonumber \\ 
& ~~~~~ - \NewMFMCBoundConstant(\policy; \dataset) - 2 \sqrt{\frac{-\ln \probOfFailure}{2 \nEpisodes}} ~\Bigg ) \Bigg | ~ \Bigg | \dataset \Bigg ] \label{eq:rademacher_estimate_mfmc}
\end{align}
where we move from Equation \ref{eq:rademacher_estimate_mfmc_1} to Equation \ref{eq:rademacher_estimate_mfmc} using 
\begin{align}
\sigma_n \sum_{t=0}^{T-1} \rewardFunction(\state_t^{n}) \geq & - \tilde{\nEpisodes} + \sigma_n \Bigg ( \estimatedReturnMFMC(\policy; \dataset) \nonumber \\ 
& ~~~~~~~~~~~ - \NewMFMCBoundConstant(\policy; \dataset) - 2 \sqrt{\frac{-\ln \probOfFailure}{2 \nEpisodes}}~ \Bigg ) \label{eq:tmp_2}
\end{align}
from Equation \ref{eq:mfmc_pac_bound_to_estimate_return} $\forall n \in \{1, ..., \tilde{N}\}$ and we assumed $-1 \leq \sum_{t=0}^{\episodeLength-1} \rewardFunction(\state_t) \leq 0$ for simplicity.

\subsection{Structures of Policy Classes}
\label{sec:structures_of_policy_classes}
In Section \ref{sec:srm_subsection} we defined a structure of function classes, from which a function class and function from that class are chosen using Equation \ref{eq:srm_objective}.
To use structural risk minimization for RL, we must similarly define a structure, 
\begin{equation}
{\policyClass_1 \subseteq \policyClass_2 \subseteq \cdots \subset \policyClass_\functionSetIndex \subseteq \cdots}, \label{eq:rl_structure}
\end{equation}
of policy classes\footnote{Note that the structure is technically over $\returnFunction \circ \policyClass_\functionSetIndex$ but ${\policyClass_i \subseteq \policyClass_j \implies \returnFunction \circ \policyClass_i \subseteq \returnFunction \circ \policyClass_j}$.}.
For RL we add the additional constraint that the policies must be Lipschitz continuous (Section \ref{sec:rl}) in order to use the bound provided in Section \ref{sec:mfmc_pac_bound}.
Note that the structure ({\it e.g.}, the indexing order $1, \dots, k, \dots$) must be specified {\it before} any data is seen.

Fortunately, some function classes have a ``natural'' ordering which may be taken advantage of, for example support vector machines \citep{vapnik98statistical} use decreasing margin size as an ordering of the structure.
In RL, many common policy representations contain natural structure and are also Lipschitz continuous.
Consider a policy class consisting of the linear combinations of radial basis functions \citep{menache05basis}.
This class is Lipschitz continuous and using this representation, we may impose a structure by progressively increasing a limit on the magnitude of all basis functions, therefore high $k$ allows for a greater range of actions a policy may choose.
This policy class consists of policies of the form 
\begin{equation}
\policy(\state) = \sum_{i=1}^M \phi_i e^{-c(\state - \bar{\state}_i)^2} \label{eq:radial_basis_fn}
\end{equation}
where $c \geq 0$,  $\bar{\state}_i \in \stateSpace$ are fixed beforehand and the progressively increasing limit $l_k \geq |\phi_i|, ~i \in \{1, \dots, M\}$ and $k$ is the index of the policy class structure.

A second policy representation which meets our requirements consists of policies of the form
\begin{equation}
\policy(\state) = \sum_{i=1}^M \frac{\phi_i}{|| \state - l_i ||_{\stateSpace}} \label{eq:our_policies}
\end{equation}
where $l_i \in \stateSpace$ are fixed beforehand and a policy of this representation is described by set $\phi_i \in \actionSpace, ~i \in \{1, \dots, M\}$.
Using this representation, we then present two possible structures, the second of which we use for the experiments in Section \ref{sec:results}.
The first places a cap on $\phi_i$, where $-a_k \leq \phi_i \leq a_k,~ \forall i \in \{1, \dots, M\}$ and $a_k \leq a_{k+1}$.
The second structure ``ties'' together $\phi_i, ~i \in \{1, \dots, M\}$ such that for $k=1$, $\phi_1 = \phi_2 = \dots = \phi_M$.
For $k=2$, we untie $\phi_1$ but still maintain that $\phi_2 = \dots = \phi_M$ and continue for the remaining $k$ such that Equation \ref{eq:rl_structure} is maintained.

Even though the representations described in this section are natural structures that can be used for many RL policy classes, the performance of each will strongly depend on the problem.
In general, choosing a policy class for a RL problem is often difficult and therefore it often must be left up to a designer.
We leave further investigation into automatically constructing structures or making the structure data-dependent ({\it e.g.}, \citet{shawetaylor96structural}) to future work.

\subsection{Finding the Best Policy from a Structure of Policy Classes}
\label{sec:finding_the_best}
Using a structure of policy classes as described in the previous section, we may now reformulate the objective of SRM into Structural Return Maximization for RL as
\begin{align}
\hat{\functionSetIndex} = & \arg \min_{\functionSetIndex} \estimatedReturnMFMC(\hat{\policy}_\functionSetIndex; \dataset) - \NewMFMCBoundConstant(\hat{\policy}_\functionSetIndex; \dataset) \nonumber \\ 
& ~~~~~~ - 2 (\lossMax-\lossMin) \sqrt{\frac{-\ln \probOfFailure}{2 \nEpisodes}} - \generalBoundConstant(\returnFunction \circ \policyClass_\functionSetIndex,\dataset, \probOfFailure) \label{eq:tmp_3}
\end{align}
\begin{equation}
\hat{\policy}_\functionSetIndex = \arg \min_{\policy \in \policyClass_\functionSetIndex} \estimatedReturnMFMC(\policy; \dataset) - \NewMFMCBoundConstant(\policy; \dataset), \label{eq:srm_rl_objective_1}
\end{equation}
where $\generalBoundConstant$ is computed using Equations \ref{eq:rademacher_bound}, \ref{eq:rademacher_estimate_bound}, and \ref{eq:rademacher_estimate_mfmc}.
Therefore, with a small batch of data, Equations \ref{eq:tmp_3} and \ref{eq:srm_rl_objective_1} will choose a small $\functionSetIndex$ and as we acquire more data $\functionSetIndex$ naturally grows.
To solve Equation \ref{eq:srm_rl_objective_1} we follow \citet{joseph13reinforcement} and use standard gradient decent with random restarts.

\section{Empirical Results}
\label{sec:results}
\citet{joseph13reinforcement} demonstrated the utility of a maximum return (MR) learner on a variety of simulated domains and a real-world problem with extremely complex dynamics.
As the results from \citet{joseph13reinforcement} show, MR performs poorly with little data and in this section we empirically demonstrate how Structural Return Maximization (SRM) remedies this shortcoming in a variety of domains.

The evaluation of SRM is done in comparison to MR due to MR being the only algorithm, to the best of our knowledge, in the RL literature which has identical assumptions to SRM.
For each experiment, the chosen approaches\footnote{See Section \ref{sec:introduction} for our discussion on the pitfalls of using model-based approaches in this setting.} choose policies from a designer-provided policy class.
To use SRM we imposed a structure on the policy class (where the original policy class was the largest, most expressive class in the structure) and the methodology from Section \ref{sec:finding_the_best} allowed SRM to select an appropriately sized policy class and policy from that class.
The MR learner maximizes the empirical return (Equation \ref{eq:empirical_return}) using the single, largest policy class.

We compare these approaches on three simulated problems: a 1D toy domain, the inverted pendulum domain, and an intruder monitoring domain.
The domains demonstrate how SRM naturally grows the policy class as more data is seen and is far less vulnerable to over-fitting with small amounts of data than a MR learner.
Policy classes were comprised of linear combinations of radial basis functions (Section \ref{sec:structures_of_policy_classes}) and training data was collected from a random policy.
To piece together artificial trajectories for MFMC we used $\tilde{\nEpisodes} = (0.1) \nEpisodes$ where $\nEpisodes$ is the number of data episodes. 

\begin{figure}[t]
\centering
\subfigure[1D Toy Domain] {
\includegraphics[width=.47\linewidth]{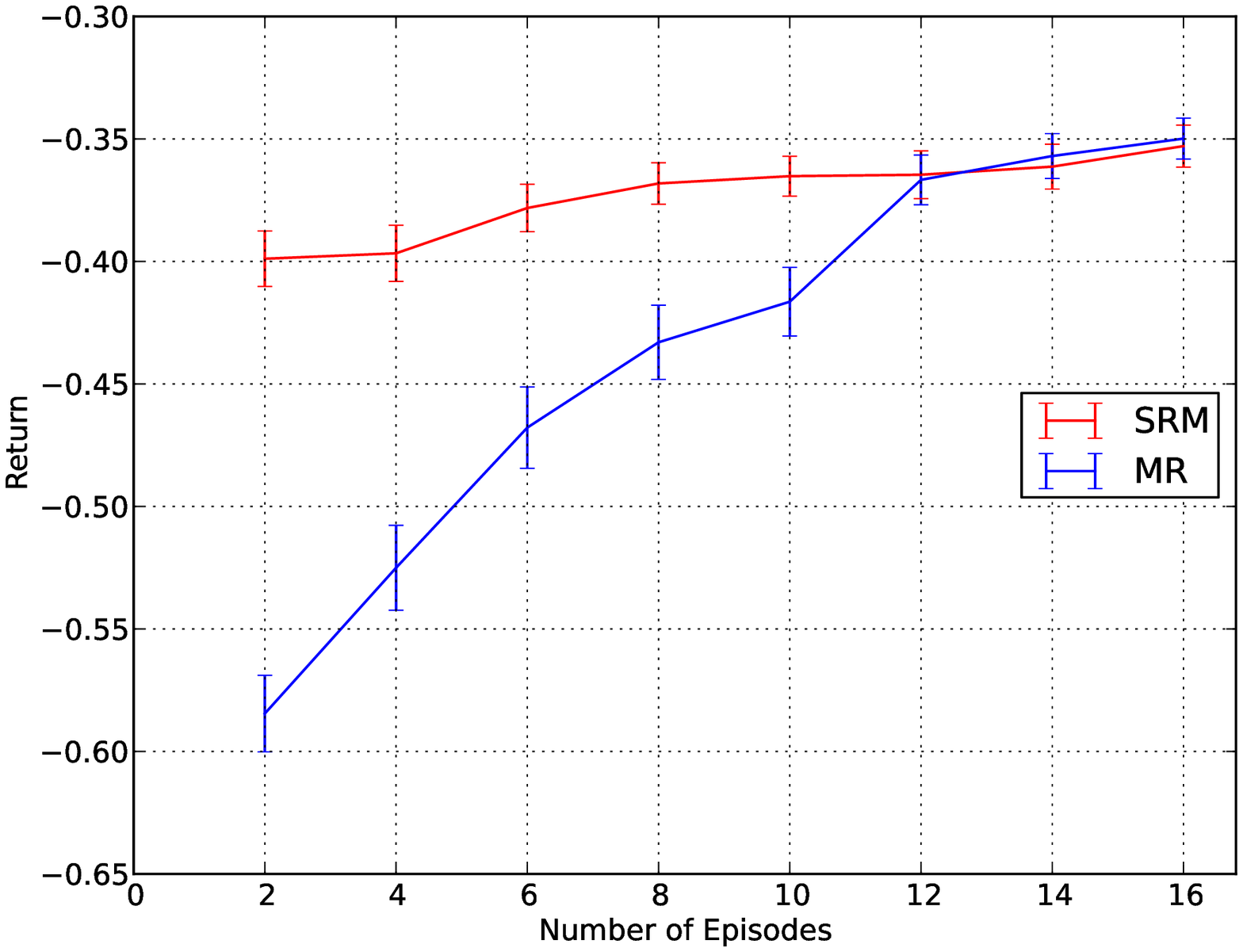}
\label{fig:flat_return_vs_data_zoomed} }
\subfigure[1D Toy Domain] {
\includegraphics[width=.47\linewidth]{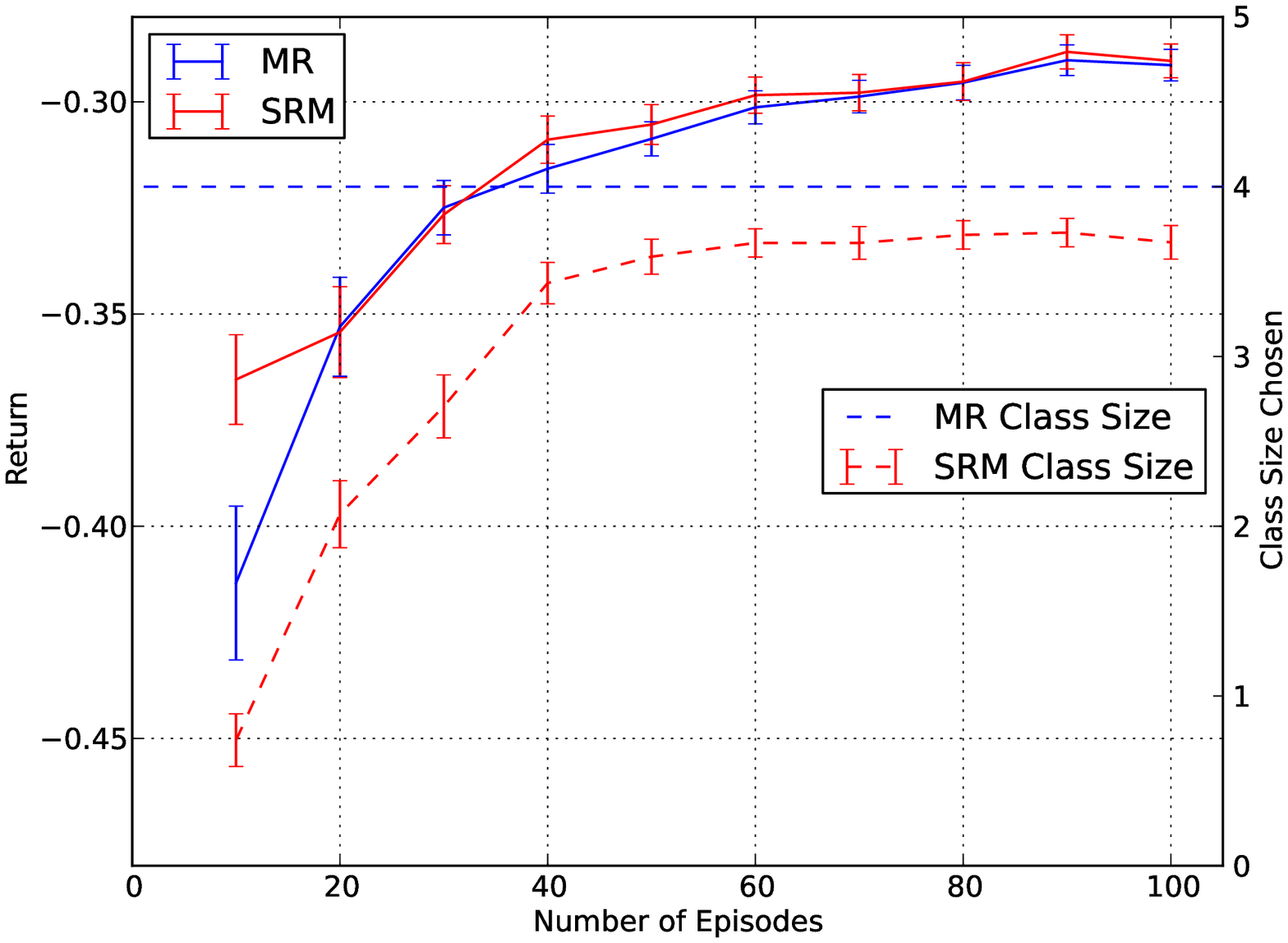}
\label{fig:flat_return_vs_data} }
\subfigure[Inverted Pendulum] {
\includegraphics[width=.47\linewidth]{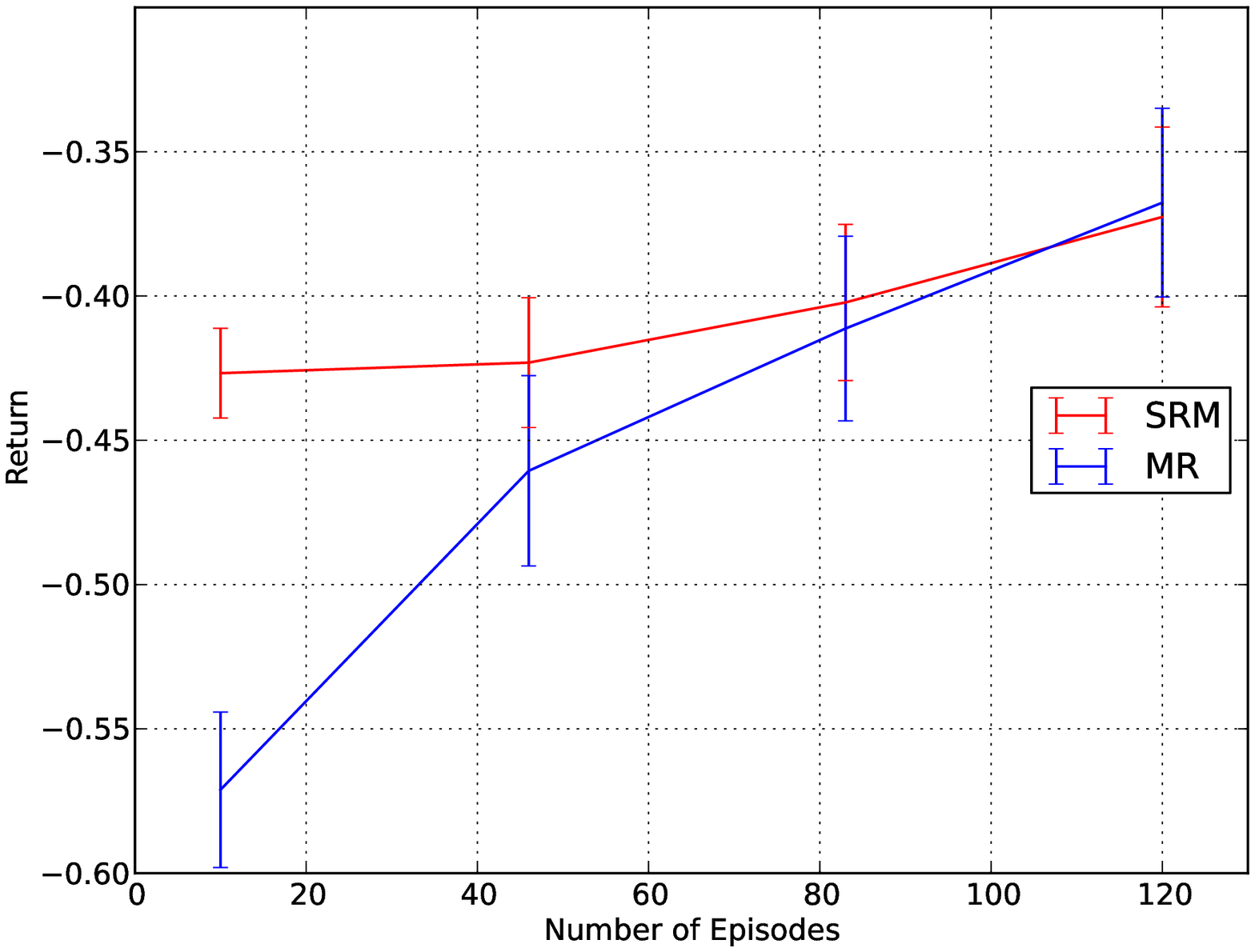}
\label{fig:inverted_pendulum_return_vs_data} }
\subfigure[Intruder Monitoring] {
\includegraphics[width=.47\linewidth]{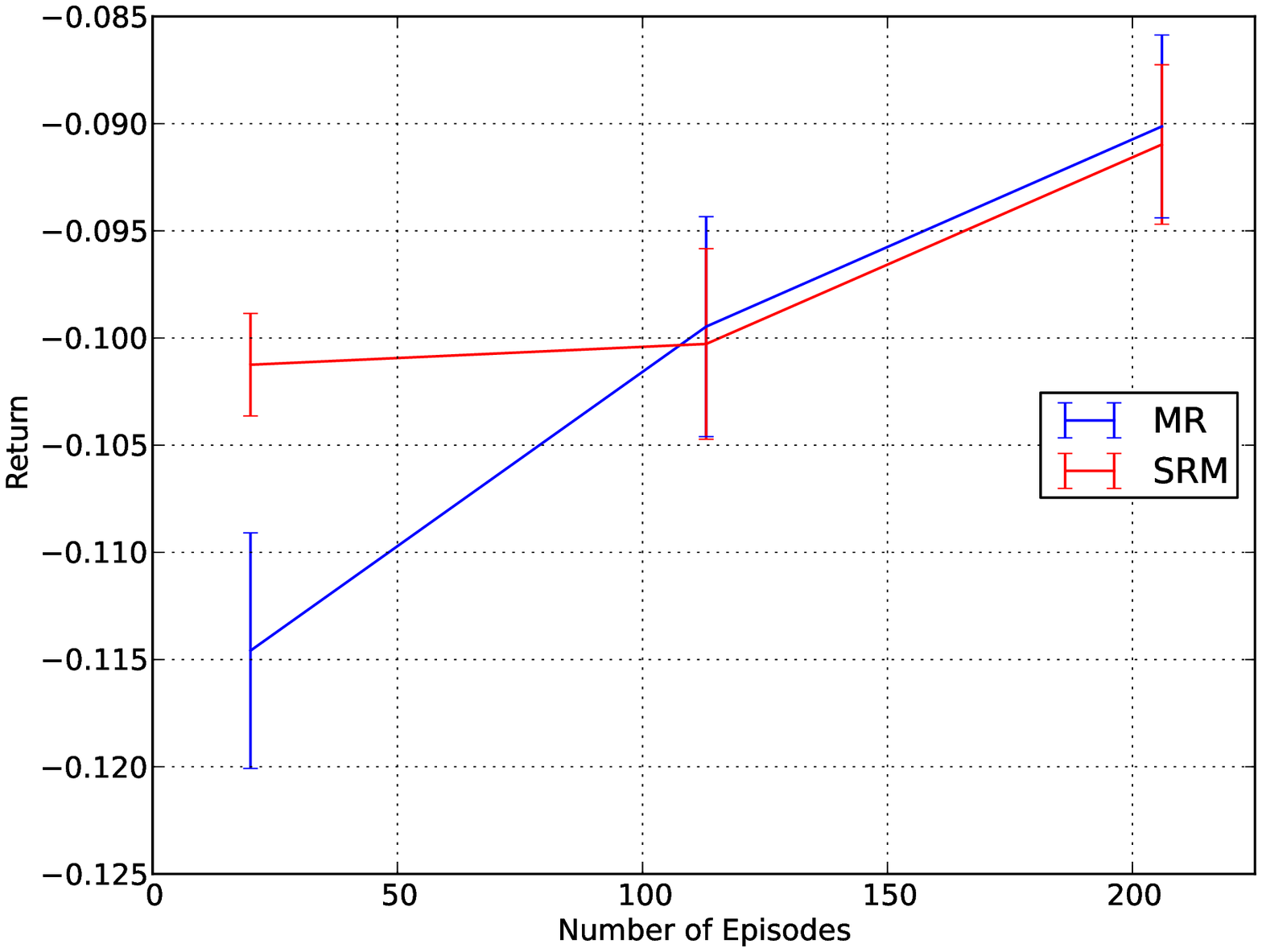}
\label{fig:intruder_return_vs_data} }
\caption{\small Performance versus the amount of training data and class size versus the amount of training data (a, b) on the 1D toy domain. Performance versus the amount of training data on the inverted pendulum domain (c) and the intruder monitoring domain (d). Error bars represent the 95\% confidence interval of the mean.}
\end{figure}

\subsection{1D Toy Domain}
\label{sec:flat}
The purpose of the 1D toy domain is to enable understanding for the reader.
The domain is a single dimensional world consisting of an agent who begins at $\state_0=0$ and attempts to ``stabilize'' at $\state=0$ in the presence of noise.
The dynamics are $\state_{t+1} = \state_t + \action + e$, $\state_t \in [-1, 1]$ and $e$ is a uniform random variable over $[-1/4, 1/4]$.
The agent takes actions $\action \in [-0.5, 0.5]$ and has reward function $\rewardFunction(\state_t) = 5 |\state_t|$.
For the policy representation we used four evenly spaced radial basis functions and for SRM we imposed five limits $l_k \in \{0, 0.125, 0.25, 0.375, 0.5\}$ on $|\phi_i|$ (see Equation \ref{eq:radial_basis_fn}).

Figures \ref{fig:flat_return_vs_data_zoomed} and \ref{fig:flat_return_vs_data} show the performance of SRM (red solid line) and MR (blue solid line) on the 1D toy domain, where figure \ref{fig:flat_return_vs_data_zoomed} is zoomed in to highlight the SRM's performance with small amounts of data.
The plot shows that MR over-fits the small amounts of data, resulting in poor performance.
On the other hand, SRM overcomes the problem of over-fitting by selecting a policy class which is appropriately sized for the amount of data.
Figure \ref{fig:flat_return_vs_data} illustrates how SRM (red dashed line) selects a larger policy class as more data is seen, in contrast to MR, which uses a fixed policy class (blue dashed line).
The figures show that as SRM is given more data, it selects increasing larger classes, allowing it learn higher performing policies without over-fitting.

\subsection{Inverted Pendulum}
\label{sec:inverted_pendulum}
The inverted pendulum is a standard RL benchmark problem (see \citet{lagoudakis03least} for a detailed explanation and parameterization of the system).
In our experiments we started the pendulum upright with the objective of learning policies which stabilize the pendulum in the presence of noise.
For the policy representation we placed 16 evenly spaced radial basis functions and for SRM we imposed two limits on $|\phi_i|$ (Equation \ref{eq:radial_basis_fn}), $l_k \in \{0, 50\}$.

Figure \ref{fig:inverted_pendulum_return_vs_data} shows the performance of SRM (red) and MR (blue) on inverted pendulum.
Similar to the results on the 1D toy domain, we see that MR over-fits the training data early on, resulting in poor performance.
In contrast, SRM achieves higher performance early on by using a small policy class with small amounts of data and growing the policy class as more data is seen.

\subsection{Intruder Monitoring}
\label{sec:intruder_monitoring}

The intruder monitoring domain models the scenario of an intruder transversing a two dimensional world where a camera must monitor the intruder around a sensitive location.
The camera observes a circle of radius $\IntruderDomainCameraRadius{} = \IntruderDomainCameraRadiusActual{}$ centered at \IntruderDomainCameraLocation{}, and the intruder, located at \IntruderDomainIntruderLocations{}, wanders toward the sensitive location with additive uniform noise.
The camera dynamics follow \IntruderDomainCameraDynamics{}, where the agent takes action $\IntruderDomainActionSymbol \in \IntruderDomainActionDomain{}$ and has reward $\IntruderDomainRewardFunction{}$ where  $\IntruderDomainBadLocationSymbol{} = \IntruderDomainBadLocationActual{}$.
For our policy representation, we placed \IntruderDomainNumStatePointsActual{} radial basis points on a grid inside \IntruderDomainWorldBounds{} and for SRM we imposed \IntruderDomainActionPointsActual{} limits on $|\RadiaBasisFunctionWeightI{}|$, $l_k \in \{0, 0.1\}$. 

Figure \ref{fig:intruder_return_vs_data} shows the performance of SRM (red) and MR (blue) on inverted pendulum.
Similar to the results on the previous domains, we see that SRM outperforms MR due to using a small policy class with small amounts of data and growing the policy class as more data is seen.

\section{Related Work}
\label{sec:related_work}

While there has been a significant amount of prior work relating Reinforcement Learning (RL) and classification \citep{langford03reducing, agoudakis03reinforcement, barto04reinforcement, langford05relating}, to the best of our knowledge, our work is the first to sufficiently develop the mapping to allow the analysis presented in Section \ref{sec:srm_classification} to be transfered to RL.
There has also been work to use classifiers to represent policies in RL \citep{bagnell03policy, rexakis08classifier, dimitrakakis08rollout, blatt05fromweighted}, which is tangential to our work; our focus is on using the principle Structural Risk Minimization for RL.
Additional work uses classification theory to bound performance for on-policy data \citep{lazaric10analysis, Farahmand12generalized}, for which Section \ref{sec:prob_mfmc_est} can be seen as extending to batch, off-policy data.

A second class of approaches aim to prevent over-fitting with small amounts of data by using both frequentist and Bayesian forms of regularization \cite{strens2000bayesian, abbeel2005exploration, bartlett2009regal, doshi2010nonparametric}.
These methods either lack formal guarantees similar to Equation \ref{eq:rl_bound} for batch data or require strong assumptions about the form of the true dynamics model, the true value function, or the optimal policy.

The RL literature also has a great deal of work growing representations as more data is seen.
Past work in the model-based \citep{doshivelez09theinfinite, joseph11abayesian} and value-based \citep{ratitch04sparse, whiteson07adaptive, geramifard11online} settings have proven successful but generally require either prior distributions or a large amount of training data collected under a specific policy.
Additionally, our approach applies in model-based, value-based, and policy search settings by treating either the model class or value function class as indirect policy representations.

\section{Conclusion}
\label{sec:conlusion}

In this work we applied Structural Risk Minimization to Reinforcement Learning (RL) to allow us to learn appropriately sized policy classes for a given amount of batch data.
The resulting algorithm had provable performance bounds under extremely weak assumptions.
To accomplish this we presented a mapping of classification to RL which allowed us to transfer the theoretical bounds previously developed in the context of classification.
These bounds allowed us to learn the policy from a structured policy class which maximized the bound on return.
We demonstrated the benefit of our approach on a 1D toy, inverted pendulum, and intruder monitoring domains as compared to an agent which naively maximizes the empirical return of the single, large policy class.

\bibliography{icml14}

\begin{thebibliography}{42}
\providecommand{\natexlab}[1]{#1}
\providecommand{\url}[1]{\texttt{#1}}
\expandafter\ifx\csname urlstyle\endcsname\relax
  \providecommand{\doi}[1]{doi: #1}\else
  \providecommand{\doi}{doi: \begingroup \urlstyle{rm}\Url}\fi

\bibitem[Abbeel \& Ng(2005)Abbeel and Ng]{abbeel2005exploration}
Abbeel, Pieter and Ng, Andrew~Y.
\newblock Exploration and apprenticeship learning in reinforcement learning.
\newblock In \emph{ICML}, 2005.

\bibitem[Anthony \& Bartlett(1999)Anthony and Bartlett]{anthony99neural}
Anthony, Martin and Bartlett, Peter~L.
\newblock \emph{Neural Network Learning: Theoretical Foundations}.
\newblock Cambridge University Press, 1999.

\bibitem[Asian et~al.(2009)Asian, Yildiz, and Alpaydin]{asian09calculating}
Asian, Ozlem, Yildiz, Olcay~Taner, and Alpaydin, Ethem.
\newblock Calculating the vc-dimension of decision trees.
\newblock In \emph{ISCIS 2009, 14-16 September 2009, North Cyprus}, pp.\
  193--198. IEEE, 2009.

\bibitem[Bagnell et~al.(2003)Bagnell, Kakade, Ng, and
  Schneider]{bagnell03policy}
Bagnell, J.~Andrew, Kakade, Sham, Ng, Andrew, and Schneider, Jeff.
\newblock Policy search by dynamic programming.
\newblock In \emph{NIPS}, 2003.

\bibitem[Bartlett \& Mendelson(2003)Bartlett and
  Mendelson]{bartlett03rademacher}
Bartlett, Peter~L. and Mendelson, Shahar.
\newblock Rademacher and gaussian complexities: risk bounds and structural
  results.
\newblock \emph{Journal of Machine Learning}, 2003.

\bibitem[Bartlett \& Tewari(2009)Bartlett and Tewari]{bartlett2009regal}
Bartlett, Peter~L and Tewari, Ambuj.
\newblock Regal: A regularization based algorithm for reinforcement learning in
  weakly communicating mdps.
\newblock In \emph{UAI}, 2009.

\bibitem[Bartlett et~al.(2002{\natexlab{a}})Bartlett, Boucheron, and
  Lugosi]{bartlett02model}
Bartlett, Peter~L., Boucheron, St{\'e}phane, and Lugosi, G{\'a}bor.
\newblock Model selection and error estimation.
\newblock \emph{Machine Learning}, 48\penalty0 (1-3):\penalty0 85--113,
  2002{\natexlab{a}}.

\bibitem[Bartlett et~al.(2002{\natexlab{b}})Bartlett, Bousquet, and
  Mendelson]{bartlett02local}
Bartlett, Peter~L., Bousquet, Olivier, and Mendelson, Shahar.
\newblock Local rademacher complexities.
\newblock In \emph{Annals of Statistics}, pp.\  44--58, 2002{\natexlab{b}}.

\bibitem[Barto \& Dietterich(2004)Barto and Dietterich]{barto04reinforcement}
Barto, A.G. and Dietterich, T.G.
\newblock Reinforcement learning and its relationship to supervised learning.
\newblock \emph{Handbook of learning and approximate dynamic programming},
  2004.

\bibitem[Baxter \& Bartlett(2001)Baxter and Bartlett]{baxter01infinite}
Baxter, Jonathan and Bartlett, Peter~L.
\newblock Infinite-horizon policy-gradient estimation.
\newblock \emph{Journal of Artificial Intelligence Research}, 2001.

\bibitem[Bertsekas(2000)]{bertsekas00dynamic}
Bertsekas, Dimitri~P.
\newblock \emph{Dynamic Programming and Optimal Control}.
\newblock Athena Scientific, 2000.

\bibitem[Blatt \& Hero(2006)Blatt and Hero]{blatt05fromweighted}
Blatt, Doron and Hero, Alfred.
\newblock From weighted classification to policy search.
\newblock In Weiss, Y., Sch\"{o}lkopf, B., and Platt, J. (eds.), \emph{NIPS}.
  2006.

\bibitem[Cherkassky \& Mulier(1998)Cherkassky and Mulier]{cherkassky98learning}
Cherkassky, Vladimir~S. and Mulier, Filip.
\newblock \emph{Learning from Data: Concepts, Theory, and Methods}.
\newblock John Wiley \& Sons, Inc., New York, NY, USA, 1st edition, 1998.

\bibitem[Dimitrakakis \& Lagoudakis(2008)Dimitrakakis and
  Lagoudakis]{dimitrakakis08rollout}
Dimitrakakis, Christos and Lagoudakis, Michail~G.
\newblock {R}ollout {S}ampling {A}pproximate {P}olicy {I}teration.
\newblock \emph{Machine Learning}, 72\penalty0 (3):\penalty0 157--171,
  September 2008.

\bibitem[Doshi-Velez(2009)]{doshivelez09theinfinite}
Doshi-Velez, Finale.
\newblock The infinite partially observable markov decision process.
\newblock In \emph{NIPS}, 2009.

\bibitem[Doshi-Velez et~al.(2010)Doshi-Velez, Wingate, Roy, and
  Tenenbaum]{doshi2010nonparametric}
Doshi-Velez, Finale, Wingate, David, Roy, Nicholas, and Tenenbaum, Joshua.
\newblock Nonparametric bayesian policy priors for reinforcement learning.
\newblock 2010.

\bibitem[Farahmand et~al.(2012)Farahmand, Precup, and
  Ghavamzadeh]{Farahmand12generalized}
Farahmand, Amir-Massoud, Precup, Doina, and Ghavamzadeh, Mohammad.
\newblock Generalized classification-based approximate policy iteration.
\newblock In \emph{European Workshop on Reinforcement Learning}, 2012.

\bibitem[Fonteneau et~al.(2010)Fonteneau, Murphy, Wehenkel, and
  Ernst]{fonteneau10model}
Fonteneau, Raphael, Murphy, Susan~A., Wehenkel, Louis, and Ernst, Damien.
\newblock Model-free monte carlo-like policy evaluation.
\newblock \emph{Journal of Machine Learning Research - Proceedings Track},
  2010.

\bibitem[Fonteneau et~al.(2012)Fonteneau, Murphy, Wehenkel, and
  Ernst]{fonteneau12batch}
Fonteneau, Raphael, Murphy, SusanA, Wehenkel, Louis, and Ernst, Damien.
\newblock {Batch mode reinforcement learning based on the synthesis of
  artificial trajectories}.
\newblock 2012.

\bibitem[Geramifard et~al.(2011)Geramifard, Doshi, Redding, Roy, and
  How]{geramifard11online}
Geramifard, Alborz, Doshi, Finale, Redding, Joshua, Roy, Nicholas, and How,
  Jonathan.
\newblock Online discovery of feature dependencies.
\newblock In \emph{ICML}, 2011.

\bibitem[Hoeffding(1963)]{hoeffding63probability}
Hoeffding, Wassily.
\newblock Probability inequalities for sums of bounded random variables.
\newblock \emph{Journal of the American Statistical Association}, 58\penalty0
  (301):\penalty0 13--30, March 1963.

\bibitem[Joseph et~al.(2011)Joseph, Doshi-Velez, Huang, and
  Roy]{joseph11abayesian}
Joseph, Joshua, Doshi-Velez, Finale, Huang, Albert~S., and Roy, Nicholas.
\newblock {A Bayesian Nonparametric Approach to Modeling Motion Patterns}.
\newblock \emph{{Autonomous Robots}}, 31\penalty0 (4):\penalty0 383--400, 2011.

\bibitem[Joseph et~al.(2013)Joseph, Geramifard, Roberts, How, and
  Roy]{joseph13reinforcement}
Joseph, Joshua, Geramifard, Alborz, Roberts, John~W., How, Jonathan~P., and
  Roy, Nicholas.
\newblock Reinforcement learning with misspecified model classes.
\newblock In \emph{ICRA}, 2013.

\bibitem[Lagoudakis \& Parr(2003{\natexlab{a}})Lagoudakis and
  Parr]{agoudakis03reinforcement}
Lagoudakis, Michail~G. and Parr, Ronald.
\newblock Reinforcement learning as classification: Leveraging modern
  classifiers.
\newblock In \emph{ICML}, 2003{\natexlab{a}}.

\bibitem[Lagoudakis \& Parr(2003{\natexlab{b}})Lagoudakis and
  Parr]{lagoudakis03least}
Lagoudakis, Michail~G. and Parr, Ronald.
\newblock Least-squares policy iteration.
\newblock \emph{JMLR}, 4:\penalty0 1107--1149, 2003{\natexlab{b}}.

\bibitem[Langford(2005)]{langford05relating}
Langford, John.
\newblock Relating reinforcement learning performance to classification
  performance.
\newblock In \emph{ICML}, 2005.

\bibitem[Langford \& Zadrozny(2003)Langford and Zadrozny]{langford03reducing}
Langford, John and Zadrozny, Bianca.
\newblock Reducing t-step reinforcement learning to classification, 2003.

\bibitem[Lazaric et~al.(2010)Lazaric, Ghavamzadeh, Munos,
  et~al.]{lazaric10analysis}
Lazaric, A., Ghavamzadeh, M., Munos, R., et~al.
\newblock Analysis of a classification-based policy iteration algorithm.
\newblock 2010.

\bibitem[Lee et~al.(1995)Lee, Bartlett, and Williamson]{lee95lower}
Lee, Wee~Sun, Bartlett, Peter, and Williamson, Robert.
\newblock Lower bounds on the vc-dimension of smoothly parametrized function
  classes.
\newblock \emph{Neural Computaion}, 1995.

\bibitem[McDonald et~al.(2011)McDonald, Shalizi, and
  Schervish]{mcdonald11estimated}
McDonald, Daniel, Shalizi, Cosma, and Schervish, Mark.
\newblock Estimated vc dimension for risk bounds.
\newblock 2011.

\bibitem[Menache et~al.(2005)Menache, Mannor, and Shimkin]{menache05basis}
Menache, Ishai, Mannor, Shie, and Shimkin, Nahum.
\newblock Basis function adaptation in temporal difference reinforcement
  learning.
\newblock \emph{Annals OR}, 134\penalty0 (1):\penalty0 215--238, 2005.

\bibitem[Meuleau et~al.(2000)Meuleau, Peshkin, Kaelbling, and
  Kim]{meuleau00offpolicy}
Meuleau, Nicolas, Peshkin, Leonid, Kaelbling, Leslie, and Kim, Kee.
\newblock Off-policy policy search.
\newblock Technical report, MIT ArtiÞcical Intelligence Laboratory, 2000.

\bibitem[Ratitch \& Precup(2004)Ratitch and Precup]{ratitch04sparse}
Ratitch, Bohdana and Precup, Doina.
\newblock Sparse distributed memories for on-line value-based reinforcement
  learning.
\newblock In \emph{Machine Learning: ECML}, Lecture Notes in Computer Science,
  2004.

\bibitem[Rexakis \& Lagoudakis(2008)Rexakis and
  Lagoudakis]{rexakis08classifier}
Rexakis, Ioannis and Lagoudakis, Michail~G.
\newblock Classifier-based policy representation.
\newblock In \emph{ICMLA}, 2008.

\bibitem[Shao et~al.(1969)Shao, Cherkassky, and Li]{shao69measuring}
Shao, Xuhui, Cherkassky, Vladimir, and Li, William.
\newblock Measuring the vc-dimension using optimized experimental design.
\newblock \emph{Neural Computation}, 12:\penalty0 2000, 1969.

\bibitem[Shawe-Taylor et~al.(1996)Shawe-Taylor, Holloway, Bartlett, Williamson,
  and Anthony]{shawetaylor96structural}
Shawe-Taylor, John, Holloway, Royal, Bartlett, Peter~L., Williamson, Robert~C.,
  and Anthony, Martin.
\newblock Structural risk minimization over data-dependent hierarchies, 1996.

\bibitem[Strens(2000)]{strens2000bayesian}
Strens, Malcolm.
\newblock A bayesian framework for reinforcement learning.
\newblock In \emph{ICML}, pp.\  943--950, 2000.

\bibitem[Sutton \& Barto(1998)Sutton and Barto]{sutton98reinforcement}
Sutton, Richard~S. and Barto, Andrew~G.
\newblock \emph{Reinforcement Learning: An Introduction}.
\newblock MIT Press, May 1998.

\bibitem[Vapnik(1998)]{vapnik98statistical}
Vapnik, Vladimir.
\newblock \emph{Statistical learning theory}.
\newblock Wiley, 1998.
\newblock ISBN 978-0-471-03003-4.

\bibitem[Vapnik et~al.(1994)Vapnik, Levin, and LeCun]{vapnik94measuring}
Vapnik, Vladimir, Levin, Esther, and LeCun, Yann.
\newblock Measuring the vc-dimension of a learning machine.
\newblock \emph{Neural Computation}, 6\penalty0 (5):\penalty0 851--876, 1994.

\bibitem[Vapnik(1995)]{vapnik95thenature}
Vapnik, Vladimir~N.
\newblock \emph{The nature of statistical learning theory}.
\newblock Springer-Verlag New York, Inc, 1995.

\bibitem[Whiteson et~al.()Whiteson, Taylor, and Stone]{whiteson07adaptive}
Whiteson, Shimon, Taylor, Matthew, and Stone, Peter.
\newblock Adaptive tile coding for value function approximation.
\newblock Technical report, University of Texas at Austin.

\end{thebibliography}
\bibliographystyle{icml2014}

\end{document}